\documentclass[letterpaper, 10 pt, conference]{ieeeconf} 

\IEEEoverridecommandlockouts
\usepackage[utf8]{inputenc}
\usepackage[T1]{fontenc}
\usepackage{amsfonts}

\usepackage{makecell}   % in preamble
\usepackage{amssymb}    % for checkmark
\usepackage{rotating} 
\usepackage{amsmath}
\usepackage{amssymb}
\usepackage{mathtools}

\usepackage{graphicx}
\usepackage{float}
\usepackage{stfloats}  % For better float positioning
\usepackage{booktabs}  % Professional table formatting
\usepackage{multirow}
\usepackage{array}
\usepackage{subcaption}
\usepackage[linesnumbered]{algorithm2e}
\usepackage{graphicx}   % for \includegraphics
\usepackage{subcaption} % for subfigures
\usepackage{textcomp}
\usepackage{blindtext}
\usepackage{lettrine}
\usepackage{lipsum}
\usepackage[font=small,skip=0pt]{caption}
\usepackage{color}
\usepackage{layouts}
\usepackage{balance}

\usepackage{url} % Ensure proper hyperlink handling
\usepackage{cite}
\usepackage{hyperref}        % Clickable links for citations
\hypersetup{
    colorlinks=true, 
    linkcolor=black, 
    citecolor=blue, 
    urlcolor=blue
}

\newcolumntype{L}[1]{>{\raggedright\let\newline\\\arraybackslash\hspace{0pt}}m{#1}}
\newcolumntype{C}[1]{>{\centering\let\newline\\\arraybackslash\hspace{0pt}}m{#1}}
\newcolumntype{R}[1]{>{\raggedleft\let\newline\\\arraybackslash\hspace{0pt}}m{#1}}
\newcolumntype{P}[1]{>{\centering\arraybackslash}p{#1}}

\usepackage[left=2cm, right=2cm, top=2cm, bottom=2.5cm]{geometry}

\title{\LARGE \bf
Robust Integrated Planning and Control for Quadrotors in Dynamic Environments via NMPC with CBF Penalties
}

\author{
Zeinab Shayan, Mohammadreza Izadi, Reza Faieghi
\thanks{*This work was partially supported by the Natural Sciences and Engineering Research Council of Canada (NSERC).}
\thanks{Z. Shayan and M. Izadi contributed equally to this paper.}
\thanks{The authors are with the Autonomous Vehicles Laboratory, Department of Aerospace Engineering, Toronto Metropolitan University, Toronto, Canada. {\tt\small \{zshayan, mizadi, reza.faieghi\}@torontomu.ca}}
}

\makeatletter
\g@addto@macro\@maketitle{
\vspace{-1em} % Adjust vertical space above the image
\begin{center}
    \centering
    \includegraphics[trim={0cm, 7.5cm, 0cm, 1cm}, clip, scale=0.5]{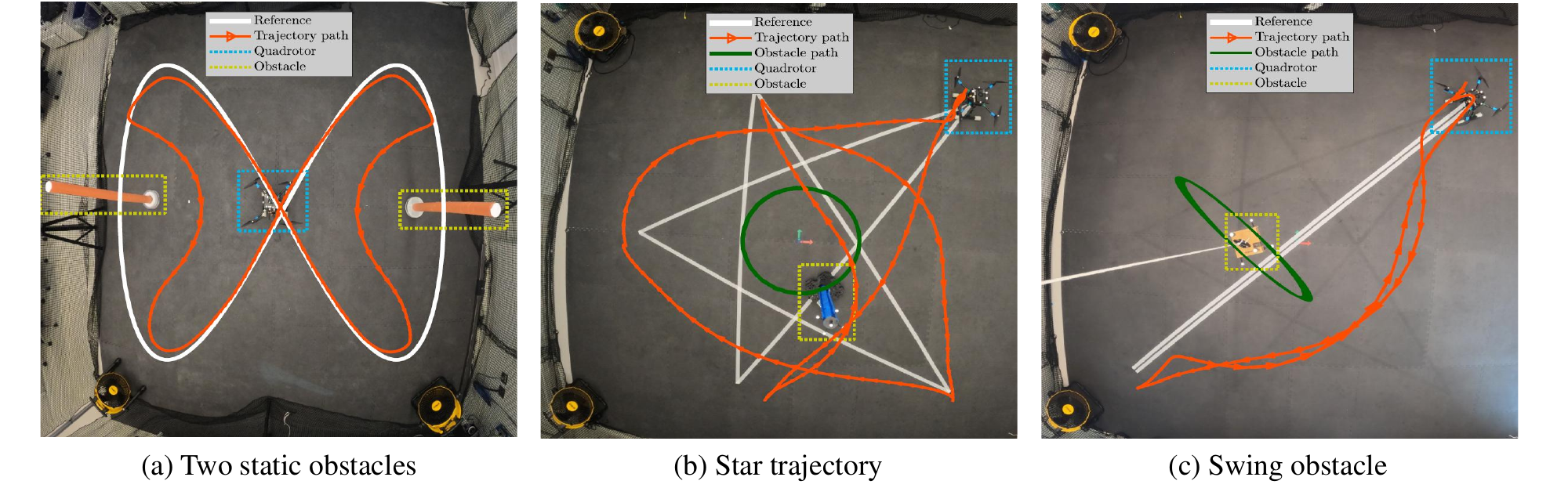}
    \captionsetup{type=figure}\setcounter{figure}{0}
    \captionof{figure}{Hardware experiment setup across different scenarios. The quadrotor follows the white reference path while avoiding: (a) two static obstacles, (b) an obstacle following a circular trajectory, and (c) a fast-moving swinging obstacle. The orange plot represents the quadrotor's trajectory.}
    \label{fig:experiment_setup}
\end{center}
\vspace{-1em} % Adjust vertical space below the image
}
\makeatother

\begin{document}

\maketitle
\thispagestyle{empty}
\pagestyle{empty}

% Set IEEE control for bibliography
\bstctlcite{IEEEexample:BSTcontrol}
\markboth{IEEE Robotics and Automation Letters}
{First Author et al.: Paper Title}

% Abstract
\begin{abstract}
This paper presents a new robust integrated planning and control (IPC) strategy for multirotor uncrewed aerial vehicles. We propose a nonlinear model predictive control (NMPC) formulation that embeds control barrier functions (CBFs) as exponential penalties, improving feasibility while ensuring smooth obstacle avoidance under tight input bounds. The penalty weights provide a practical tuning knob to trade off tracking accuracy against avoidance aggressiveness. We enhance the system robustness by employing a high-gain disturbance observer (HGDO) to estimate and compensate for external disturbances. We also incorporate a Kalman filter (KF) for computationally efficient, real-time prediction of obstacle motion, enabling avoidance of moving obstacles. Comparative studies against both conventional NMPC and NMPC with hard CBF constraints, validated in Gazebo and hardware experiments, demonstrate superior feasibility, safety, and robustness. To the best of our knowledge, this is the first hardware-validated NMPC–CBF IPC framework, offering a practical step toward safe quadrotor deployment in dynamic environments.
\end{abstract}

% Sections
\section{Introduction}
\subsection{Related Work and Motivation}
Safe navigation of quadrotors in cluttered and dynamic environments requires reactive motion planning algorithms that enable rapid, real-time responses to environmental changes and external disturbances. 

One reactive planning approach that has received significant attention in recent years is Model Predictive Control (MPC).
% , which accounts for system dynamics as well as input and state constraints. It
It can handle both static and dynamic obstacles by incorporating either hard constraints \cite{luis2020online, castillo2018model, ahn2022model, toumieh2024high, shayan2025nonlinear} or penalty terms into the cost function \cite{bui2024model, small2019aerial, kamel2018review}. However, as shown in \cite{shayan2025exponential}, the conventional approach of defining constraints or penalty terms based solely on the relative position of the vehicle and obstacles do not guarantee safety.

To address this limitation, an alternative approach is the use of Control Barrier Functions (CBFs) \cite{ames2016control, ames2014control, nguyen2016exponential}. By using the Lie derivatives of the CBF along system trajectories, CBFs account for the vehicle-obstacle dynamics, not just their relative position in a given time. Integrating CBFs into MPC thus combines the predictive power of MPC with the safety guarantees of CBFs.

A few studies have explored MPC-CBF integration for motion planning of multirotor uncrewed aerial vehicles (MRUAVs) \cite{ali2024mpc, wang2024dual, goarin2024decentralized, shayan2025exponential}. These studies have treated planning and control as separate layers. The planner generates collision-free reference trajectories, while the controller ensures tracking. As a result, planning does not consider disturbance rejection, and control does not explicitly account for obstacle avoidance. This limitation becomes critical in agile flight, where deviations from the planned trajectory increase the risk of collision, since obstacles in the updated trajectory are not considered. Additionally, when an obstacle appears suddenly, the control layer alone may not be able to ensure safety. 

% Specifically, \cite{ali2024mpc} proposed a cascaded MPC-CBF scheme with dynamic feedback linearization, while \cite{wang2024dual} introduced a dual MPC framework integrating a Control Lyapunov Function (CLF) for formation control and a CBF for collision avoidance. However, both approaches remain limited to simulations and static obstacles. Moreover, \cite{shayan2025exponential} used CBFs as constraints for an MPC-based planner that can be integrated with common autopilot programs such as PX4. 

Integrated Planning and Control (IPC) aims to address the above challenge of separating planning and control \cite{liu2023integrated, romero2022model, minavrik2024model, wu2021external}. 
A few existing IPC studies consider a linearized quadrotor model \cite{liu2023integrated, ahn2022model, toumieh2024high}. However, this limits the flight envelope given the high nonlinearity of quadrotor dynamics.
A few other existing IPC works use Nonlinear MPC (NMPC) to enable more agility and robustness \cite{romero2022model, minavrik2024model, wu2021external}, but have not explored obstacle avoidance or their results have been limited to static obstacles.

% Polyhedral representations are particularly effective for box-shaped or planar obstacles; however, they are limited in representing irregular or non-convex free spaces. In environments with narrow passages or intricate geometries, these constraints can become overly conservative, reducing available flight paths and potentially resulting in suboptimal or infeasible solutions. Compared to polyhedral representations, ellipsoids \cite{luis2020online, castillo2018model, wu2021external} are particularly well-suited for representing irregular or rounded obstacles, where they can provide tighter and more compact bounding volumes. Their smooth, differentiable boundaries also make them amenable to integration with gradient based optimization and CBF frameworks.

Overall, the existing IPC studies for MRUAVs \cite{romero2022model, minavrik2024model, wu2021external, liu2023integrated} exhibit several limitations.
First, their obstacle avoidance strategies rely solely on position-based constraints or penalty terms, which provide weaker safety guarantees compared to CBF-based strategies.
Second, most do not incorporate obstacle motion predictions into the MPC framework, limiting their effectiveness against moving obstacles.
Third, several existing IPC approaches do not explicitly account for external disturbances \cite{romero2022model, minavrik2024model, liu2023integrated}, compromising robustness.

\subsection{Contributions}
In this paper, we present a new IPC framework that combines NMPC and CBF while effectively handling external disturbances and moving obstacles, addressing the aforementioned limitations of prior IPC studies.

At its core, our IPC framework leverages the complementary strengths of NMPC and CBF. In addition, we incorporate a high-gain disturbance observer (HGDO) \cite{izadi2024hgdo} for real-time estimation and compensation of external disturbances, and a Kalman filter (KF) to predict obstacle trajectories, which we embed into the NMPC prediction horizon, giving rise to a prescient NMPC formulation \cite{tajeddin2019ecological}.

As detailed in our extensive Gazebo simulations and hardware experiments, our IPC framework enables robust trajectory tracking under wind disturbances while consistently ensuring safe avoidance of both static and moving obstacles in confined environments across diverse scenarios.

A central contribution of our work lies in the way we integrate CBFs within NMPC. Conventionally, CBFs are imposed as hard constraints in the NMPC formulation. However, prior work \cite{xiao2021adaptive} has shown that simultaneously enforcing CBFs and actuator limits as hard constraints can drastically reduce the feasible solution set for the optimization problem, often leading to infeasibility and subsequent instability. 

A potential mitigation is to apply Adaptive CBFs \cite{xiao2021adaptive}, but this comes at the cost of additional states for each obstacle, extensive parameter tuning, and increased computational overhead, especially for systems with high relative degree such as MRUAVs. 

However, we adopt a different strategy: we embed the CBF directly in the NMPC stage cost as smooth exponential cost penalties. 
As it is shown in our experiments, especially Figs.~\ref{fig:static_2d}--\ref{fig:u_gazebo_lem}, this strategy recasts safety constraints into differentiable penalties that preserve feasibility, reduce oscillatory behavior compared to using a CBF term as a constraint, and retain the dynamics-aware shaping induced by CBFs. 

Moreover, the penalty weights provide an effective tuning knob to trade tracking accuracy against avoidance aggressiveness. While this soft-constraint design relaxes hard safety guarantees, we recover practical safety by sharply increasing the penalty as the system approaches constraint boundaries and by selecting margins that account for model mismatch and disturbances.

\section{PRELIMINARIES}
\subsection{Notations}
Throughout this paper, unless stated otherwise, we use the following notation standards.
The two coordinate frames used in this paper include the inertial frame $\mathcal{F}_\mathcal{I} = \{\mathbf{x}_\mathcal{I}, \mathbf{y}_\mathcal{I}, \mathbf{z}_\mathcal{I}\}$, configured in an East-North-Up (ENU) orientation, and the body frame $\mathcal{F}_\mathcal{B} = \{\mathbf{x}_\mathcal{B}, \mathbf{y}_\mathcal{B}, \mathbf{z}_\mathcal{B}\}$, fixed at the quadrotor’s center of mass and aligned in a Forward-Left-Up (FLU) orientation (Fig. \ref{fig:frame}).
We denote the $\ell_2$-norm of $\mathbf{v}$ as $\|\mathbf{v}\|$. We represent the dot product between two vectors $\mathbf{u}$ and $\mathbf{v}$ as $\mathbf{u}\cdot\mathbf{v}$, and their cross product as $\mathbf{u}\times\mathbf{v}$. 
We express unit quaternions as $\mathbf{q} = [q_w,\vec{\mathbf{q}}^\top]^\top\in \mathbb{S}^3$, where  $\mathbb{S}^3$ refers to the unit 3-sphere. We denote the multiplication of
two unit quaternions $\mathbf{p}$ and $\mathbf{q}$ by $\mathbf{p}\otimes\mathbf{q}$. For a unit quaternion
$\mathbf{q}$, we define an operator $B_\mathbf{q}:\mathbb{R}^3 \to \mathbb{R}^3$ as
\begin{equation}
B_\mathbf{q}(\mathbf{u}) = \left(q_w^2 - \|\vec{\mathbf{q}}\|^2\right) \mathbf{u} + 2 (\vec{\mathbf{q}} \cdot \mathbf{u}) \vec{\mathbf{q}} + 2 q_w (\vec{\mathbf{q}} \times \mathbf{u}).
\end{equation}
\subsection{Quadrotor dynamics}
Let ${\boldsymbol{\xi}}=\left[x,y,z\right]^\top$ denote the vehicle position, $\mathbf{q} = [q_w,\vec{\mathbf{q}}^\top]^\top$ the attitude, $\boldsymbol{\upsilon}=\left[u,v,w\right]^\top$ the linear velocity, and ${\boldsymbol{\omega}}=\left[p,q,r\right]^\top$ the angular velocity. The quadrotor dynamic equations are expressed by: 
\begin{figure}[t]
    \centering
    \includegraphics[trim={10cm, 8cm, 14cm, 4.2cm}, clip, width = 0.7\linewidth]{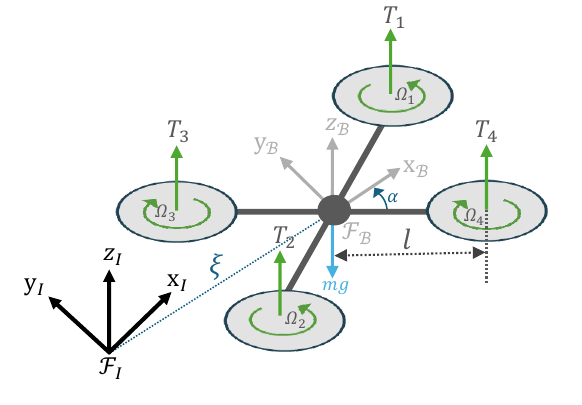}
    \caption{A schematic diagram of quadrotor and Coordinate frames}
    \label{fig:frame}
\end{figure}
\begin{equation}
\label{eq:quadrotor_model}
\begin{aligned}
    \left\{
    \begin{array}{l}
    \dot{\boldsymbol{\xi}} = \boldsymbol{\upsilon}, \\
    \dot{\boldsymbol{\upsilon}} = -\mathbf{g} + \frac{1}{m} B_\mathbf{q} (\mathbf{f}_{\mathbf{u}}) +  \frac{1}{m} \mathbf{f}_{\mathbf{d}}, \\
    \dot{\mathbf{q}} = \frac{1}{2} \mathbf{q} \otimes [0,\boldsymbol{\omega}^\top]^\top,\\
    \dot{\boldsymbol{\omega}} = \mathbf{J}^{-1} \left(  - \boldsymbol{\omega} \times \mathbf{J} \boldsymbol{\omega} + \boldsymbol{\tau}_{\mathbf{u}} + \boldsymbol{\tau}_{\mathbf{d}} \right).
    \end{array}
    \right.
\end{aligned}
\end{equation}
where $m$ is the mass of the vehicle, ${\bf{J}} = {\rm{diag}}\left(J_{xx},J_{yy},J_{zz}\right)$ is the inertia matrix, ${{\bf{g}}} = \left[0,0,g\right]^\top$ is the gravity vector with $g$ set to $9.81\;[m/s^2]$, ${\bf{f}_d}$ and ${\boldsymbol{\tau}}_{\mathbf{d}}$ are the thrust and moments caused by disturbances, and ${\bf{f}}_{\mathbf{u}}=\left[0,0,f_c\right]^\top$ and ${\boldsymbol{\tau}}_{\mathbf{u}}=\left[\tau_x,\tau_y,\tau_z\right]^\top$ are the thrust and moments caused by rotors. Each rotor produces a thrust and a torque as
\begin{equation}\label{eq:thrust}
T_i=k_t\Omega_i^2,\tau_i=k_q\Omega_i^2,
\end{equation}
where $k_t$ and $k_q$ are the rotors' thrust and torque coefficients, and $\Omega_i$ is the rotor angular speed.
Then the collective thrust magnitude ${f_c}$ and ${\boldsymbol{\tau}_{\mathbf{u}}}$ can be expressed as follows
\begin{equation}\label{eq:f}
\left[{f_c, \boldsymbol{\tau}_{\mathbf{u}}}\right]^\top = \mathbf{H}\left[ T_1,T_2,T_3,T_4 \right]^\top,
\end{equation}
where  $\mathbf{H}$ is
\begin{equation}\label{eq:tau}
\mathbf{H} =
\begin{bmatrix}
1 & 1 & 1 & 1 \\
- l \sin \alpha & l \sin \alpha & - l \sin \alpha & l \sin \alpha \\
- l \cos \alpha & l \cos \alpha & - l \cos \alpha & l \cos \alpha \\
- k_q / k_t & - k_q / k_t & k_q / k_t & k_q / k_t
\end{bmatrix},
\end{equation}
and $l$ and $\alpha$ are the parameters defined in Fig. \ref{fig:frame}.

\section{The proposed IPC architecture}
\begin{figure}[t]
    \centering
    \includegraphics[trim={7.6cm, 3cm, 5cm, 3.5cm}, clip, width = \linewidth]{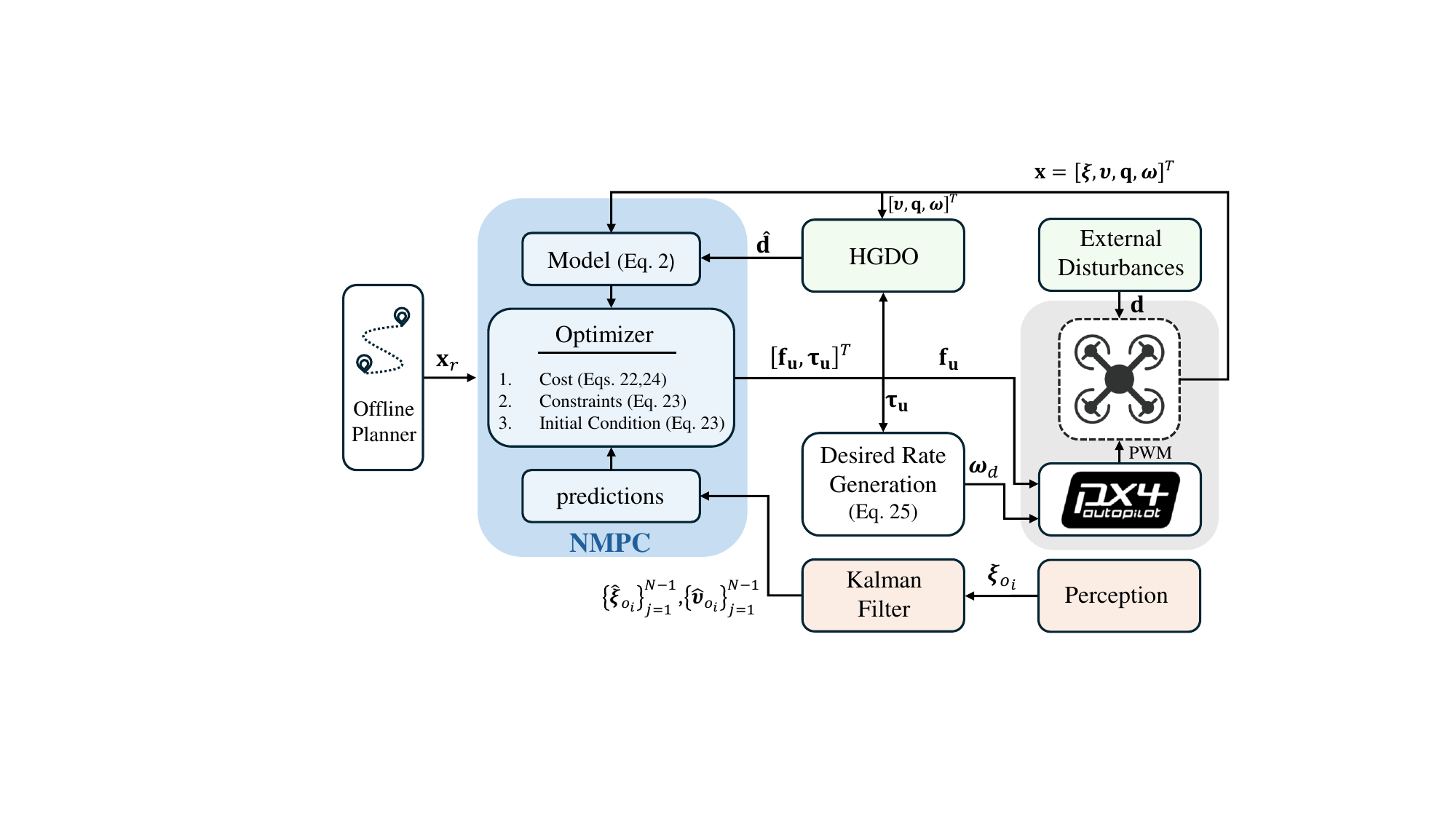}
    \caption{Block diagram of the proposed robust IPC framework using NMPC and CBF}
    \label{fig:block_diagram}
\end{figure}

Figure \ref{fig:block_diagram} outlines the proposed robust IPC architecture. 
We assume access to a dynamically feasible reference trajectory $\mathbf{x}_r$, which can be generated offline, e.g., a minimum-jerk trajectory passing through waypoints selected manually or computed by a path-finding algorithm such as $\text{A}^\star$, given a map of the environment.

In the absence of obstacles, the NMPC generates thrust and moment commands $({\mathbf f}_{\mathbf u},{\boldsymbol{\tau}}_{\mathbf u})$ to track $\mathbf{x}_r$. In the vicinity of obstacles, the CBF penalty terms \eqref{eq:cbf_cost} dominate the NMPC cost, prompting re-planning and simultaneous updates to $({\mathbf f}_{\mathbf u},{\boldsymbol{\tau}}_{\mathbf u})$ to track the new trajectory.

Obstacle detection and their positions $\boldsymbol{\xi}_{o_i}$ are provided by an external perception module. For moving obstacles, a KF predicts the $i$-th obstacle’s position $\hat{\boldsymbol{\xi}}_{o_i}$ and velocity $\hat{\boldsymbol{\upsilon}}_{o_i}$ over the prediction horizon $N$, and these predictions are used in computing the CBF penalty \eqref{eq:cbf_cost}.

To ensure robustness against external disturbances, the NMPC formulation incorporates a HGDO to estimate and compensate for disturbances in real time.

\subsection{Representing safety margins for obstacles}\label{sec_obs}
We model both the vehicle and all obstacles as axis-aligned bounding ellipsoids. An ellipsoid aligned with the inertial-frame axes is defined by
\begin{equation}
    \mathcal{E}(\mathbf{c},\mathbf{S}) = \left\{ \boldsymbol{\chi} \in \mathbb{R}^3 \mid (\boldsymbol{\chi} - \mathbf{c})^\top \mathbf{S}^{-1} (\boldsymbol{\chi} - \mathbf{c}) \leq 1 \right\},
\end{equation}
where $\mathbf{c}$ denotes the centroid of $\mathcal{E}$, and $\mathbf{S} = \text{diag}(a,b,c)$ characterizes its shape, with $a$, $b$, and $c$ representing the semi-principal axis lengths of the ellipsoid.
To distinguish between the vehicle and obstacles, we use the indices $q$ and $o_i$, referring to the bounding ellipsoids and their associated parameters for the quadrotor and the $i$-th obstacle, respectively.
For the quadrotor, we employ two concentric bounding ellipsoids: $\mathbf{S}_{q_{\min}}$, which denotes the shape matrix of the quadrotor’s tightest bounding ellipsoid, and $\mathbf{S}_s$ which introduces a precautionary safety margin. We then define $\mathbf{S}_q = \mathbf{S}_{q_{\min}} + \mathbf{S}_s$.

The ellipsoidal separation condition, which can serve as a safety constraint within the conventional NMPC framework, is expressed as follows
\begin{equation}\label{eq:ellipsoidal_separation_condition} (\boldsymbol{\xi} - \boldsymbol{\xi}_{o_i})^\top \mathbf{Q}_\mathcal{E}^{-1} (\boldsymbol{\xi} - \boldsymbol{\xi}_{o_i}) -1 \geq 0,
\end{equation}
where  $\mathbf{Q}_\mathcal{E} = (\mathbf{S}_q + \mathbf{S}_{o_i})^2$.
Constraints similar to \eqref{eq:ellipsoidal_separation_condition} have been previously used in NMPC frameworks, but they do not explicitly account for vehicle-obstacle dynamics, and therefore may fail to guarantee safety in practice, particularly with dynamic obstacles, as shown in our experiments.

We enhance these constraints using CBFs. However, enforcing CBFs as hard constraints within the NMPC optimization problem, when planning and control are integrated, can cause infeasibility due to conflicts with actuator limits \cite{xiao2021adaptive}. In particular, the CBF safety constraint \eqref{eq:cbf_constraint} is input-dependent and may require control values that violate the upper/lower bounds imposed in the optimization, leading to constraint conflicts \cite{xiao2021adaptive}.
% Although the integration of CBF constraints with MPC has been studied for more than a decade, most prior work has focused primarily on the planning layer or on linear models of vehicles and obstacles. When the control layer and nonlinear models are also considered, satisfying the physical limitations of actuators in highly nonlinear problems becomes critical, and conflicts often arise. 
To address this challenge, we formulate CBF as an exponential penalty term, which improves feasibility while accounting for vehicle-obstacle dynamics.
\subsection{CBF penalty}\label{sec_cbf}
Let us first establish the CBF definition \cite{ames2014control, ames2016control}.
Consider a nonlinear system with the state $\mathbf{x} \in \mathbb{R}^n$, control input $\mathbf{u} \in \mathbb{R}^m$, and output $\mathbf{y} \in \mathbb{R}^m$ as
\begin{equation}\label{eq:generic_nonlinear_system}
    \dot{\mathbf{x}} = {\mathbf{f}}({\mathbf{x}}, {\mathbf{u}}), \quad {\mathbf{y}} = {\mathbf{g}}(\mathbf{x}),
\end{equation}
If $h({\mathbf{x}}): \mathbb{R}^n \to \mathbb{R}$ is a continuously differentiable scalar function, then the safe set $\mathcal{C} \in \mathbb{R}^n$, and  boundary and interior of $\mathcal{C}$ are defined as
\begin{align}
    \mathcal{C} &= \{ {\mathbf{x}} \in \mathbb{R}^n \mid h({\mathbf{x}}) \geq 0 \}, \\
    \partial \mathcal{C} &= \{ {\mathbf{x}} \in \mathbb{R}^n \mid h({\mathbf{x}}) = 0 \}, \\
    \text{Int}(\mathcal{C}) &= \{ {\mathbf{x}} \in \mathbb{R}^n \mid h({\mathbf{x}}) > 0 \}.
\end{align}
The set $\mathcal{C}$ is forward invariant if, for any ${\mathbf{x}}(0)=\mathbf{x}_0 \in \mathcal{C}$, the solution ${\mathbf{x}}(t) \in \mathcal{C}$ for all $t \geq 0$. Forward invariance ensures that the system remains within the safe set indefinitely.

Therefore, the constraint $h(\mathbf{x}) \ge 0$ determines $\mathbf{x} \in \mathcal{C}$; however, to establish the forward invariance of $\mathcal{C}$, we explore the evolution of $h(\mathbf{x})$ along the system trajectories, which is described by the Lie derivatives
\begin{equation}
\begin{array}{l}
    L_{\mathbf{f}}^i h({\mathbf{x}}, {\mathbf{u}}) = \frac{\partial L_{\mathbf{f}}^{i-1} h({\mathbf{x}}, {\mathbf{u}})}{\partial {\mathbf{x}}} \mathbf{f}({\mathbf{x}}, {\mathbf{u}}),
\end{array}
\end{equation}
where $i$ represents the order of the Lie derivative.  $h({\mathbf{x}})$ has a relative degree of $r > 0$ if
$\frac{\partial}{\partial \mathbf{u}} L_{\mathbf{f}}^i h({\mathbf{x}}, {\mathbf{u}}) = \mathbf{0}$ for $i = 0, \dots, r-1$, and $\frac{\partial}{\partial \mathbf{u}} L_{\mathbf{f}}^{r} h({\mathbf{x}}, {\mathbf{u}}) \neq \mathbf{0}$.

Then, $h(\mathbf{x})$ is called an exponential CBF if there exists $\mathbf{k} = \begin{bmatrix} k_0 & k_1 & \cdots & k_{r-1} \end{bmatrix}$ such that for all $\mathbf{x} \in \mathcal{C}$ 
\begin{equation}\label{eq:cbf_condition_1}
    \inf_{\mathbf{u} \in \mathcal{U}} \left[ L_{\mathbf{f}}^{r} h(\mathbf{x}, \mathbf{u}) + \mathbf{k} \boldsymbol{\zeta}(\mathbf{x}) \right] \geq 0,
\end{equation}
\begin{equation}\label{eq:cbf_condition_2}
   h(\mathbf{x}(t)) \ge C e^{({{\mathbf{F}}-{\mathbf{G}}{\mathbf{k}}})t} \boldsymbol{\zeta}(\mathbf{x}_0) \ge 0 \quad \text{when} \quad h(\mathbf{x}_0) \ge 0,
\end{equation}
where $\mathcal{U}$ represents the admissible control input set, $C=[1 \ 0 \ \cdots \ 0]$, and
\begin{equation}\label{eq:xi_dynamics}
    \dot{\boldsymbol{\zeta}} =
  \underbrace{
    \begin{bmatrix}
        0 & 1 & 0 & \cdots & 0 \\
        0 & 0 & 1 & \cdots & 0 \\
        \vdots & \vdots & \vdots & \ddots & \vdots \\
        0 & 0 & 0 & \cdots & 1 \\
        0 & 0 & 0 & \cdots & 0
    \end{bmatrix}
    }_{\mathbf{F}}
    \underbrace{
    \begin{bmatrix}
        h(\mathbf{x}) \\
        L_{\mathbf{f}} h(\mathbf{x}) \\
        L_{\mathbf{f}}^2 h(\mathbf{x}) \\
        \vdots \\
        L_{\mathbf{f}}^{r-1} h(\mathbf{x})
    \end{bmatrix}
    }_{\boldsymbol{\zeta}}
    +
    \underbrace{
    \begin{bmatrix}
        0 \\
        0 \\
        \vdots \\
        0 \\
        1
    \end{bmatrix}
    }_{\mathbf{G}}
    L_{\mathbf{f}}^{r}h(\mathbf{x}, \mathbf{u}).
\end{equation}
According to \cite{nguyen2016exponential}, the conditions \eqref{eq:cbf_condition_1}--\eqref{eq:cbf_condition_2} are equivalent to the differential inequality $\dot{\boldsymbol{\zeta}} \ge \left(\mathbf{F}-\mathbf{G}\mathbf{k}\right) \boldsymbol{\zeta}$. If $\mathbf{k}$ is chosen such that $\mathbf{F}-\mathbf{G}\mathbf{k}$ is Hurwitz, then the last row of $\dot{\boldsymbol{\zeta}} \ge \left(\mathbf{F}-\mathbf{G}\mathbf{k}\right) \boldsymbol{\zeta}$ yields
\begin{equation}\label{eq:xi_dynamics_cbf_constraint}
        L_\mathbf{f}^{r}h(\mathbf{x}) + k_{r-1}L_\mathbf{f}^{r-1}h(\mathbf{x}) + \dots + k_0 h(\mathbf{x})  \ge 0.
\end{equation}
Consequently, this ensures that the conditions \eqref{eq:cbf_condition_1} and \eqref{eq:cbf_condition_2} are satisfied, establishing $h(\mathbf{x})$ is an exponential CBF.
These inequalities guarantee that a feasible control input $\mathbf{u} \in \mathcal{U}$ directs the system to maintain safety, with the gain vector $\mathbf{k}$  regulating the rate of exponential convergence of  $h(\mathbf{x})$ toward the safe set boundary or deeper into it.

With this in mind, we define the CBF $h_{i}(\mathbf{x})$ with a relative degree of $r=2$ and its Lie derivatives with respect to the quadrotor dynamics \eqref{eq:quadrotor_model} for the $i$-th obstacle as follows
\begin{align}
   h_{i}(\mathbf{x}) &=  (\boldsymbol{\xi} - \boldsymbol{\xi}_{o_i})^\top \mathbf{Q}_{\mathcal{E}_i}^{-1} (\boldsymbol{\xi} - \boldsymbol{\xi}_{o_i}) -1, \label{eq:our_cbf} \\
   L_\mathbf{f} h_{i}(\mathbf{x}) &= 2 \Delta \boldsymbol{\xi}_{i}^\top \mathbf{Q}_{\mathcal{E}_i}^{-1} \Delta \boldsymbol{\upsilon}_i, \label{eq:first_lie_derivative} \\
   L_\mathbf{f}^2 h_{i}(\mathbf{x}) &= 2 \left( \Delta \boldsymbol{\upsilon}_i^\top \mathbf{Q}_{\mathcal{E}_i}^{-1} \Delta \boldsymbol{\upsilon}_i + \Delta \boldsymbol{\xi}_i^\top \mathbf{Q}_{\mathcal{E}_i}^{-1} \dot{\boldsymbol{\upsilon}} \right). \label{eq:second_lie_derivative}
\end{align}
where $\Delta \boldsymbol{\xi}_{i} = \boldsymbol{\xi} - \boldsymbol{\xi}_{o_i}$, $\Delta \boldsymbol{\upsilon}_{i} = \boldsymbol{\upsilon} - \boldsymbol{\upsilon}_{o_i}$, and $\boldsymbol{\xi}_{o_i}$ and $\boldsymbol{\upsilon}_{o_i}$ represent the position and velocity of the $i$-th obstacle, respectively.
As illustrated in Fig. \ref{fig:block_diagram}, at time $t$, the perception module measures $\boldsymbol{\xi}_{o_i}(t)$, and then $\boldsymbol{\xi}_{o_i}(t)$ and $\boldsymbol{\upsilon}_{o_i}(t)$ are subsequently predicted over the horizon $(t, t+T]$ using the KF. We assume that obstacles have zero acceleration and jerk within this horizon.

Now, we can ensure that the quadrotor states remain within the safe set defined by $h_i(\mathbf{x}) \ge 0$ by imposing
\begin{equation}\label{eq:cbf_constraint}
\Phi_i(\mathbf{x}, \mathbf{u}) = L_\mathbf{f}^2 h_i(\mathbf{x}) + k_1 L_\mathbf{f} h_i(\mathbf{x}) + k_0 h_i(\mathbf{x}) \geq 0.
\end{equation}

Since both the quadrotor and obstacle models are nonlinear, and planning and control are integrated, the entire IPC framework is fully nonlinear. As a result, if the control input bounds are very tight, defining \eqref{eq:cbf_constraint} as a constraint in the NMPC framework may not always yield a feasible solution. Therefore, inspired by \cite{tajeddin2019ecological}, we introduce the following exponential penalty term \eqref{eq:cbf_constraint}
\begin{equation}\label{eq:cbf_cost} 
 \mathcal{J}_i^{CBF}(\mathbf{x}, \mathbf{u}) = w_1 e^{-w_2 \Phi_i(\mathbf{x}, \mathbf{u})},
\end{equation}
where $w_1$ and $w_2$ are the tuning weights. 

The above expression \eqref{eq:cbf_cost} introduces a safety term whose value varies exponentially with the proximity to the safe set boundary. As the quadrotor approaches an obstacle, $\Phi_i(\mathbf{x}, \mathbf{u})$ decreases significantly, causing the exponential term multiplied by $w_1$ to grow rapidly. This gives the safety term a dominant term in the cost function, making safety the highest-priority objective in the optimization. Conversely, when the quadrotor is far from an obstacle, $\Phi_i(\mathbf{x}, \mathbf{u})$ takes a large positive value, and the exponential term ($e^{-w_2 \Phi_i(\mathbf{x}, \mathbf{u})} \to 0$) becomes negligible. 
% At this point, the safety concern is resolved, and the controller shifts focus to tracking objectives.

\subsection{Disturbance observer}
We employ a disturbance observer for disturbance estimation and compensation. To this end, we adopt the HGDO formulation proposed in \cite{izadi2024hgdo}. For its integration into the NMPC framework, the system model used by the NMPC replaces the disturbance term ${\mathbf{d}}=\left[{\bf{f}_{\mathbf{d}}, \boldsymbol{\tau}_{\mathbf{d}}}\right]^\top$ in \eqref{eq:quadrotor_model} with its estimate $\hat{{\mathbf{d}}} = \left[{\hat{\bf{f}}_{\mathbf{d}}, \hat{\boldsymbol{\tau}}_{\mathbf{d}}}\right]^\top$. Therefore, the NMPC control actions will account for the estimated disturbance.
\subsection{NMPC formulation}
Let us begin by defining the state vector ${\bf x} = \left[{\boldsymbol{\xi}}^\top,{\boldsymbol{\upsilon}}^\top,{\mathbf{q}}^\top,{\boldsymbol{\omega}}^\top\right]^\top$.
The goal is to design the control signal $\mathbf{u}=\left[{\bf{f}_{\mathbf{u}}, \boldsymbol{\tau}_{\mathbf{u}}}\right]^\top$ to minimize the tracking error $\bf{e} = \bf{x}-\bf{x}_r$ while avoiding obstacles. To this end, we adopt the NMPC formulation as
\begin{equation}\label{eq:mpc_formulation_1}
    \begin{split}
        \mathbf{u}^\star(t) = \min_{\mathbf{u}(\cdot)}\ \mathcal{M}\left(\mathbf{e}(t_k+T)\right) + &  \int_{t_k}^{t_k+T} \Big(\mathcal{L}\left(\mathbf{e}(t), \mathbf{u}(t)\right)  
          \Big)dt,
    \end{split}
\end{equation}
subject to
\begin{equation}
\dot{\mathbf{x}} = {\mathbf{f}}({\mathbf{x}}, {\mathbf{u}} +\hat{\mathbf{d}}), \quad \mathbf{x}(t_k) = \mathbf{x}_k,
\end{equation}
\begin{equation}
\mathbf{x}(t) \in \mathcal{X}, \quad \mathbf{u}(t) \in \mathcal{U}, \quad \forall t \in [t_k, t_k + T],
\label{eq:mpc_formulation_3}
\end{equation}
where $t_k$ is the current time, $T = N T_s$ is the prediction time, and $T_s$ is the sampling time. The function $\mathbf{f}(\mathbf{x}, \mathbf{u}, \hat{\mathbf{d}})$ represents the nonlinear dynamics defined in \eqref{eq:quadrotor_model} under disturbance, while $\hat{\mathbf{d}}$ denotes the disturbance estimates obtained from the HGDO. The sets $\mathcal{X}$ and $\mathcal{U}$ represent the admissible state and input spaces, respectively. $\mathcal{L}\left(\mathbf{e}, \mathbf{u}\right)$ corresponds to the stage cost, and $\mathcal{M}\left(\mathbf{e}\left(t_k+T\right)\right)$ represents the terminal cost, defined as
\begin{equation}\label{eq:cost_function}
\mathcal{L}(\mathbf{e}, \mathbf{u}) = \mathbf{e}^\top \mathbf{Q}_\mathbf{e} \mathbf{e} + \mathbf{u}^\top \mathbf{Q}_\mathbf{u} \mathbf{u} + \sum_{i=1}^{n} \mathcal{J}_i^{CBF}\left(\mathbf{x}, \mathbf{u}\right),
\end{equation}
\begin{equation}\label{eq:terminal_cost}
\mathcal{M}(\mathbf{e}(t_k+T)) = 0,
\end{equation}
where $\mathbf{Q}_\mathbf{e} \ge 0$ and $\mathbf{Q}_\mathbf{u} > 0$ are the weight matrices, $n$ is the number of obstacles, and $\mathcal{J}_i^{CBF}(\mathbf{x}, \mathbf{u})$ is defined by \eqref{eq:cbf_cost}.

$\mathcal{J}_i^{CBF}\left(\mathbf{x}, \mathbf{u}\right)$ dominates the stage cost in the vicinity of obstacles, prompting the optimization problem to re-plan the desired trajectory to ensure obstacle avoidance and safety. 
% In the absences of obstacles, $\mathcal{J}_i^{CBF}\left(\mathbf{x}, \mathbf{u}\right) \rig0$, and the optimization focuses solely on trajectory tracking.

Once the control input $\mathbf{u} = [\mathbf{f}_{\mathbf{u}}, \boldsymbol{\tau}_{\mathbf{u}}]^\top$ is obtained from the NMPC, the desired body rates $\boldsymbol{\omega}_d$ are computed from $\boldsymbol{\tau}_{\mathbf{u}}$ as
\begin{equation}
\label{eq:desired_rate}
\begin{array}{c}
 \qquad  \quad  \dot{\boldsymbol{\omega}}_d = \mathbf{J}^{-1} \left(\boldsymbol{\tau}_{\mathbf{u}} - \boldsymbol{\omega} \times \mathbf{J} \boldsymbol{\omega}\right),\\
\boldsymbol{\omega}_d = \boldsymbol{\omega} + T_s \dot{\boldsymbol{\omega}}_d.
\end{array}
\end{equation}
Then, $\boldsymbol{\omega_d}$ and $\bf{f}_u$ are sent to the autopilot, e.g., PX4.

To solve \eqref{eq:mpc_formulation_1}--\eqref{eq:mpc_formulation_3} in real time, we utilize the ACADOS toolkit \cite{verschueren2022acados} with the partial-condensing-HPIPM solver.
To this end, the continuous-time formulation \eqref{eq:mpc_formulation_1}--\eqref{eq:mpc_formulation_3} is discretized using a fourth-order Runge-Kutta integration scheme. $T$ is divided into $N$ (prediction horizon) intervals of equal duration $T_s$ (sampling time), such that $T = N T_s$.

\subsection{Feasibility and stability analysis}\label{sec:feasibility_stability}
The feasibility and stability properties of the NMPC formulation \eqref{eq:mpc_formulation_1}–\eqref{eq:mpc_formulation_3} align with established results in the soft-constrained NMPC literature \cite{Zeilinger2014,he2015stability}, and a full analysis is therefore omitted here due to space limitations.

We emphasize, however, that the adoption of exponential penalty terms \eqref{eq:cbf_cost}, while less conventional, is consistent with prior work demonstrating their effectiveness in preserving feasibility compared to hard constraints \cite{tajeddin2019ecological}. By construction, soft penalties relax strict feasibility requirements and generally improve solver robustness, while maintaining practical safety by sharply increasing the cost as the system approaches constraint boundaries. This behavior is confirmed in our experiments, where CBF penalties yield feasible results, whereas hard CBF constraints frequently lead to infeasibility.

Regarding stability, the NMPC formulation guarantees ultimate boundedness of the closed-loop trajectories \cite{Zeilinger2014}. Since the controller operates with disturbance estimates provided by an HGDO, the boundedness results in \cite{izadi2024hgdo} apply directly, ensuring that the closed-loop error remains uniformly ultimately bounded, with the ultimate bound adjustable via the HGDO tuning gains. Taken together, these arguments mirror the formal analyses in \cite{Zeilinger2014,izadi2024hgdo}, and we omit detailed proofs here.

\subsection{Obstacle movement predictions}\label{se:KF}
To predict obstacle motion, we model each obstacle as a double integrator and use a discrete-time KF to estimate its position and velocity.

At time $t_k$, the KF takes the centroid position $\boldsymbol{\xi}_{o_i}(t_k)$ of $\mathcal{E}_{o_i}$, measured by an external perception module, and predicts its future states over the NMPC horizon $N$
\begin{equation}
    \left\{\hat{\boldsymbol{\xi}}_{o_i}(t_k+jT_s)\right\}_{j=1}^{N-1}, \quad \left\{\hat{\boldsymbol{\upsilon}}_{o_i}(t_k+jT_s)\right\}_{j=1}^{N-1},
\end{equation}
where $\hat{\boldsymbol{\xi}}_{o_i}$ and $\hat{\boldsymbol{\upsilon}}_{o_i}$ denote the predicted position and velocity of the $i$-th obstacle. These predictions are then used to compute \eqref{eq:our_cbf}, \eqref{eq:first_lie_derivative}, and \eqref{eq:second_lie_derivative} within the horizon.

It is worth noting that the KF is chosen for its low computational cost, enabling efficient real-time prediction. While alternative methods exist, identifying the most effective approach is beyond the scope of this paper.

\section{Results and Discussion}\label{se:results}
We compared different IPC frameworks, such as conventional NMPC with position-only constraints \eqref{eq:ellipsoidal_separation_condition}, NMPC with CBF constraints \eqref{eq:cbf_constraint}, and our proposed method, NMPC with a CBF penalty cost \eqref{eq:cbf_cost}, across different scenarios. 
% This comparison demonstrates the effectiveness of NMPC with a CBF penalty cost in improving feasibility and smoothness, while also maintaining safety and tracking.

We first conducted simulations in Gazebo and then hardware experiments using a customized Holybro X500 quadrotor equipped with a Pixhawk flight controller and an onboard companion computer (Intel Core i5-1340P). The NMPC was implemented in Robot Operating System (ROS) on the companion computer, and its commands were sent to PX4 via MAVROS at 100 [Hz].

The experiments were conducted in an indoor environment equipped with an OptiTrack motion capture system, which provided real-time positions of the quadrotor and obstacles. The flight space, with dimensions $[-2.4,2.5]\times[-2.3,2.5]\times[0,3]$ $[m^3]$, was incorporated as state constraints in the NMPC formulation \eqref{eq:mpc_formulation_3}.

The parameters used in the experiments are as follows: $\alpha=45^\circ$, $l=0.25$ [m], $N=80$, $T_s=0.01$ [s], $m=2.3$ [kg], ${\bf{J}} = {\rm{diag}}\left(0.03,0.03,0.06\right)$, ${{\bf{Q}_e}} = 10^3{\rm{diag}}\left( { 8,8,10,6,6,4.5,1,1,1,1,2,2,2} \right)$, ${{\bf{Q}_u}}$ = $ 100\mathbf{I}_4$, $k_0=5$, $k_1=10$, $w_1=3000$, $w_2=0.4$. The allowable control inputs within NMPC, defined as a constraint, are set to [0,30] for thrust and [-3,3] for moments. 
We also set ${\mathbf{q}}_r = \left[ 1, 0, 0, 0 \right]^\top$, ${\boldsymbol{\omega}}_r = \left[ 0, 0, 0 \right]^\top$.

We considered three different scenarios, as illustrated in Fig. \ref{fig:experiment_setup} and described below.
\subsubsection*{Scenario 1 -- Static Obstacles Scenario} In this scenario, the quadrotor is required to follow a lemniscate trajectory, defined as ${\boldsymbol{\xi}}_r = \left[1.5\sin \left(\frac{{\pi t}}{{20}}\right),1.5\sin \left(\frac{{2\pi t}}{{20}}\right), 1 \right]^\top$, while avoiding two identical static obstacles. These obstacles are bounded by ellipsoids centered at $(1.5,0,1)^\top$ and $(-1.5,0,1)^\top$, each defined by the shape matrix $\mathbf{S}_{o}=\text{diag}(0.1, 0.1, 1)$. 

\subsubsection*{Scenario 2 -- Star trajectory with a circular moving obstacle scenario}
This scenario involves a dynamic obstacle, defined as a circularly moving ellipsoid with a shape matrix of $\mathbf{S}_{o}=\text{diag}(0.1,0.1,0.6)$. Mounted on a TurtleBot3, the obstacle follows a circular path of radius $0.5 [\text{m}]$ around the origin. The quadrotor must track a star-shaped trajectory while actively avoiding collisions.

\subsubsection*{ Scenario 3 -- Swinging obstacle scenario}
This scenario includes a swinging box as a fast-moving obstacle oscillating in a pendulum-like motion. This makes the scenario particularly challenging, as the oscillating motion continuously changes the collision zone with changing velocities and accelerations. The quadrotor should follow a straight trajectory from $(1.5,1.5,1.3)^\top$ to $(-1.5,-1.5,1.3)^\top$ and back while avoiding the obstacle.

\subsection{Gazebo Simulation Results}
% \begin{figure}[t]
%     \centering
%     \begin{subfigure}[t]{0.49\columnwidth}
%         \centering
%         \includegraphics[trim={5.5cm, 9.5cm, 6cm, 10cm}, clip, width = \linewidth]{Figures/lemniscate_traj_gazebo.pdf}
%         \caption{Static obstacles scenario}
%         \label{fig:static_2d}
%     \end{subfigure}
%     \begin{subfigure}[t]{0.49\columnwidth}
%         \centering
%         \includegraphics[trim={5.5cm, 9.5cm, 6cm, 10cm}, clip, width = \linewidth]{Figures/star_traj_gazebo.pdf}
%         \caption{Star trajectory scenario}
%         \label{fig:star_2d}
%     \end{subfigure}
%     \caption{2D view of quadrotor trajectory under different IPC frameworks across different scenarios in Gazebo 
%     % \textbf{Note the instability of NMPC + CBF constraint case in Fig. 4(b).
%     }
%     \label{fig:gazebo_traj}
% \end{figure}

\begin{figure}[t]
        \centering
        \includegraphics[trim={5.5cm, 9.5cm, 6cm, 10cm}, clip, width = 0.7\linewidth]{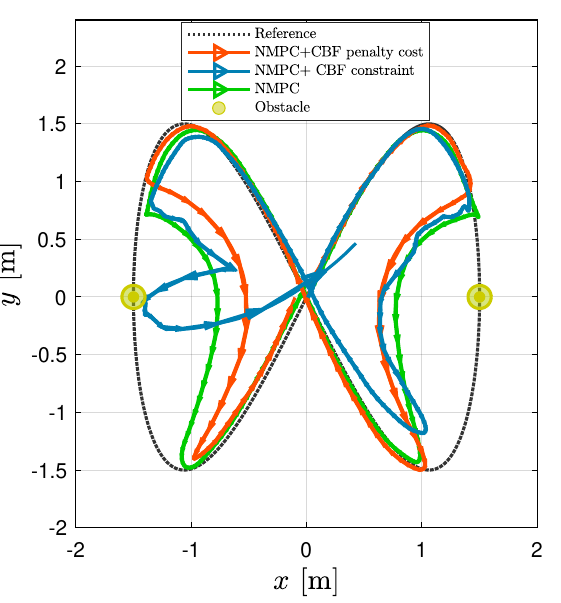}
        \caption{2D view of the quadrotor trajectory under different IPC frameworks in the static obstacles scenario (Gazebo)}
        \label{fig:static_2d}
% \end{figure}
% \begin{figure}
        \centering
        \includegraphics[trim={5.5cm, 9.5cm, 6cm, 10cm}, clip, width = 0.7\linewidth]{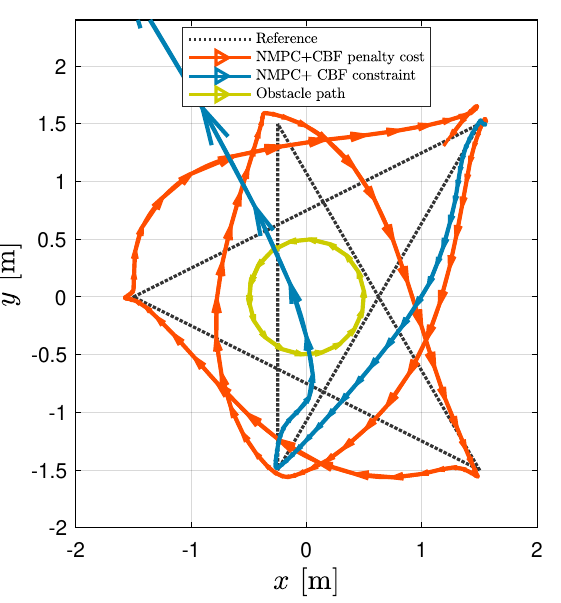}
        \caption{2D view of the quadrotor trajectory under different IPC frameworks in the star-trajectory scenario (Gazebo)}
        \label{fig:star_2d}
    % \caption{2D view of quadrotor trajectory under different IPC frameworks across different scenarios in Gazebo 
    % \textbf{Note the instability of NMPC + CBF constraint case in Fig. 4(b).
    
    % \label{fig:gazebo_traj}
\end{figure}

\begin{figure}[t]
    \centering
    \includegraphics[trim={4.5cm, 10.5cm, 4.5cm, 11.2cm}, clip, width = \linewidth]{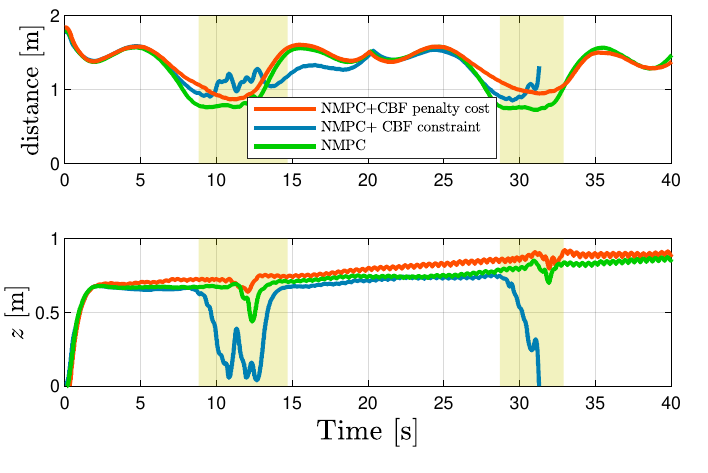}
    \caption{Minimum quadrotor-obstacle distance and altitude $z$ in the static-obstacle scenario (Gazebo). Yellow shading indicates obstacle-encounter episodes.}
    \label{fig:dis_gazebo_lem}
% \end{figure}
% \begin{figure}[t]
    \centering
    \includegraphics[trim={4.5cm, 7.5cm, 4.5cm, 8cm}, clip, width = \linewidth]{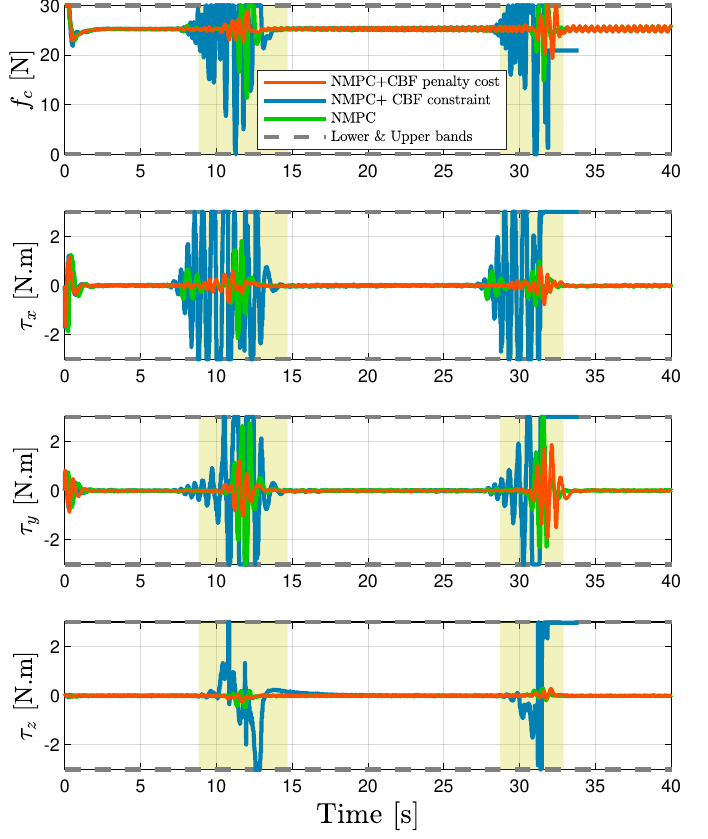}
    \caption{Control inputs under different IPC frameworks in the static-obstacle scenario (Gazebo). Yellow shading indicates obstacle-encounter episodes.
    }
    \label{fig:u_gazebo_lem}
    \centering
\end{figure}
Figures~\ref{fig:static_2d}--\ref{fig:star_2d} compare the quadrotor’s trajectories under different IPC frameworks in different scenarios. In the case of static obstacles, NMPC produces a smooth, collision-free path. However, in the presence of moving obstacles this results in collisions because it relies only on position constraints. As a result, we omitted the NMPC trajectory from Fig. \ref{fig:star_2d}.

With NMPC augmented by CBF constraints, the vehicle initially avoids the first static obstacle in Fig. \ref{fig:static_2d}, but then develops growing oscillations that lead to infeasibility and instability mid-mission. The same instability appears in Fig. \ref{fig:star_2d}. In contrast, our proposed NMPC with CBF penalty yields significantly smoother, safer, and feasible navigation in both static and dynamic settings.

For a more detailed discussion on the static scenario, let us consider Figs.~\ref{fig:dis_gazebo_lem} and \ref{fig:u_gazebo_lem}. The NMPC+CBF constraint exhibits growing oscillations as the vehicle nears the obstacle. In Fig.~\ref{fig:dis_gazebo_lem}, around $t=9$ [s], the altitude $z$ drops sharply and the vehicle briefly lands to preserve safety. After $t=9$ [s] it takes off again but continues oscillating until it clears the obstacle and rejoins the reference path. After $t=29$ [s], however, the quadrotor cannot proceed: the solver reports infeasibility at $t=31.45$ [s].

As shown in Fig. \ref{fig:u_gazebo_lem}, the NMPC+CBF constraint drives the control inputs into repeated saturation, producing large fluctuations. This induces oscillatory motion and frequent instability, indicating that the input bounds are too tight to satisfy the CBF constraints concurrently.

By contrast, the other two IPC frameworks generate smooth, safe trajectories: control inputs, altitude, and minimum obstacle distance remain well behaved, with the NMPC+CBF penalty formulation achieving a larger safety margin than the baseline NMPC (Tab.~\ref{tab:nmpc_variants}).

The feasibility challenge of the NMPC+CBF constraint approach becomes more pronounced when accounting for dynamic obstacles. As shown in Tab.~\ref{tab:nmpc_variants}, in the star scenario, the solver becomes infeasible much earlier in the mission, around $t = 12.42$ [s]. 

Despite partial success with the NMPC+CBF constraint formulation in Gazebo, all hardware trials failed. In contrast, NMPC+CBF penalty achieved collision-free flight in all hardware tests, across static and dynamic obstacles and under external disturbances, maintaining a consistent safety margin and eliminating oscillations and infeasibility, as detailed in the next section.

\begin{table}[t]
\centering
\caption{Performance comparison of different IPC frameworks across different scenarios in Gazebo}
\label{tab:nmpc_variants}
\begin{tabular}{@{}l@{\hspace{8mm}}l@{\hspace{8mm}}l@{\hspace{8mm}}l@{}}\toprule
 & NMPC & \begin{tabular}[c]{@{}l@{}}NMPC+CBF\\ Constraint\end{tabular} & \begin{tabular}[c]{@{}l@{}}NMPC+CBF\\ Penalty\end{tabular} \\ \midrule
\multicolumn{4}{l}{\textbf{Static scenario:}} \\
Outcome & Successful & \begin{tabular}[c]{@{}l@{}}Infeasible \\ at $t=31.45$ {[}s{]}\end{tabular} & Successful \\
\begin{tabular}[c]{@{}l@{}}Min. obstacle\\ clearance\end{tabular} & 0.7271 {[}m{]} & 0.8548 {[}m{]} & 0.8698 {[}m{]} \\ \midrule
\multicolumn{4}{l}{\textbf{Star scenario:}} \\
Outcome & Collision & \begin{tabular}[c]{@{}l@{}}Infeasible\\ at $t=12.42$ {[}s{]}\end{tabular} & Successful \\
\begin{tabular}[c]{@{}l@{}}Min. obstacle\\ clearance\end{tabular} & - & - & 0.6804 [m] \\ \bottomrule
\end{tabular}
\end{table}
\subsection{Hardware Experiments}
\subsubsection{Evaluation without external disturbance}
Figure \ref{fig:experiment_setup} presents the experimental setup for all scenarios. The plotted trajectories correspond to the no-disturbance case. The proposed NMPC+CBF penalty tracks the reference and avoids obstacles, capabilities that the NMPC+CBF constraint could not realize due to infeasibility. The swinging obstacle is the most challenging case, with fast, nonlinear motion; nevertheless, integrating KF for obstacle motion prediction helps achieve collision-free results.

\begin{figure}[t]
    \centering
    \includegraphics[trim={4cm, 12cm, 4.8cm, 12.3cm}, clip, width = \linewidth]{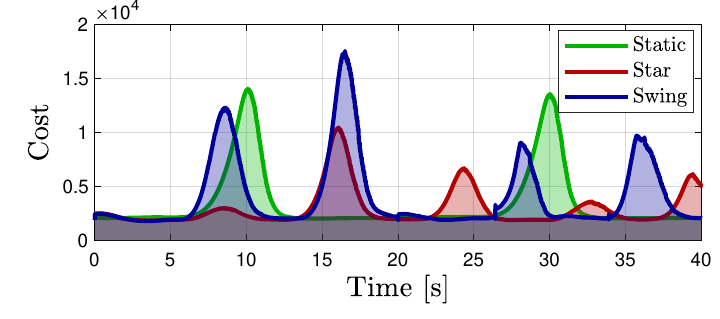}
    \caption{Stage cost value for the proposed NMPC+CBF penalty method across different scenarios (hardware experiments)}
    \label{fig:cost_real}
% \end{figure}
% \begin{figure}[t]
    \centering
    \includegraphics[trim={4cm, 12cm, 4.8cm, 12.5cm}, clip, width = \linewidth]{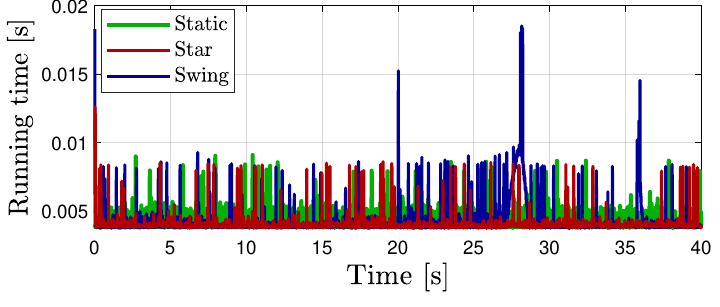}
    \caption{IPC running time for the proposed NMPC+CBF penalty method across different scenarios (hardware experiments)}
    \label{fig:solver_real}
% \end{figure}
% \begin{figure}[t]
    \centering
    \includegraphics[trim={5.3cm, 9.5cm, 6cm, 10.3cm}, clip, width =0.7 \linewidth]{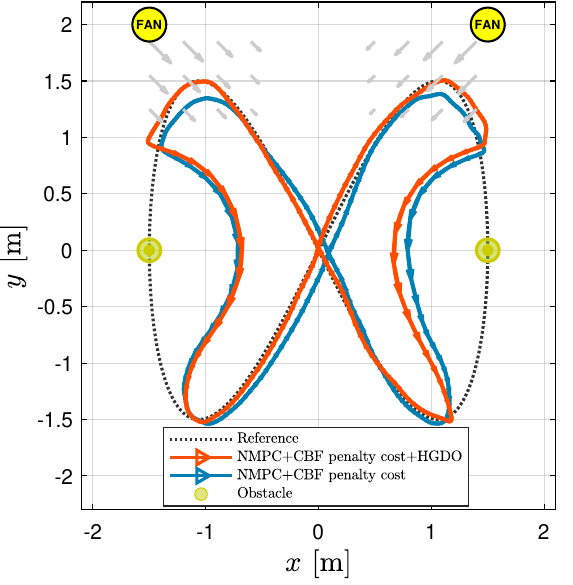}
    \caption{2D view of the quadrotor trajectory in the static-obstacle scenario under disturbance (hardware experiments).}
    \label{fig:traj_real}
\end{figure}

As shown in Fig. \ref{fig:cost_real}, the NMPC stage cost rises as the quadrotor approaches an obstacle and then drops sharply as it recedes, demonstrating our method’s ability to achieve trajectory tracking and obstacle avoidance simultaneously. 

To assess computational load, we measured per-scenario runtimes (Fig. \ref{fig:solver_real}), which remained consistently below 0.02 [s]. This fits within the 20 [ms] budget of PX4’s 50 [Hz] position-control loop, satisfying real-time planning and control requirements for many applications.

\subsubsection{Evaluation in the presence of disturbance}
\begin{figure}[t]
    \centering
    \includegraphics[trim={4cm, 7.5cm, 4cm, 8cm}, clip, width = \linewidth]{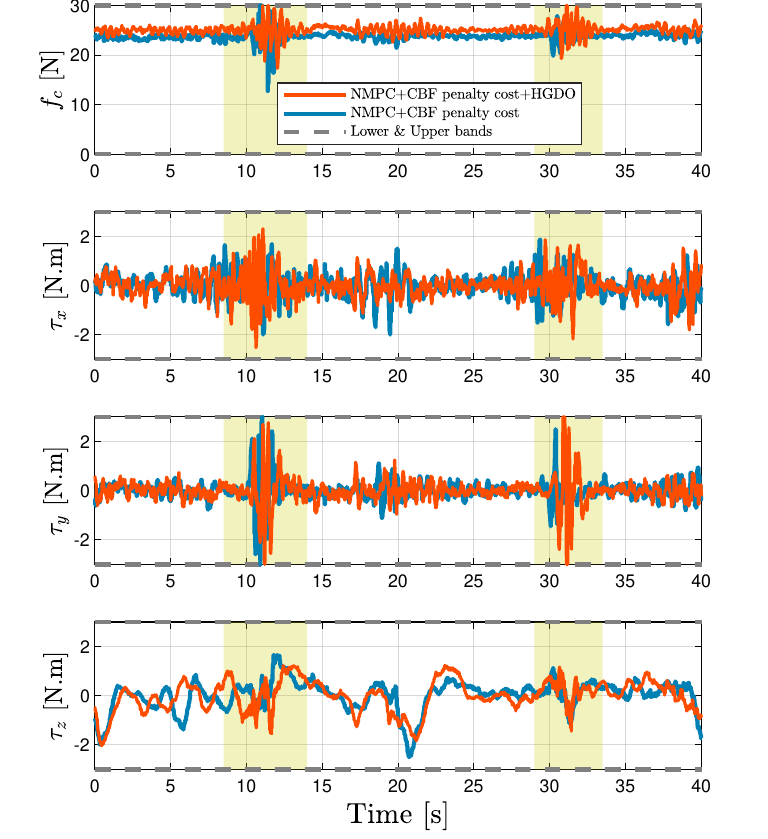}
    \caption{Control signals in the static-obstacle scenario under disturbance (hardware experiments)}
    \label{fig:u_real}
    \centering
% \end{figure}
% \begin{figure}[t]
    \centering
    \includegraphics[trim={4cm, 10.5cm, 4cm, 11cm}, clip, width = \linewidth]{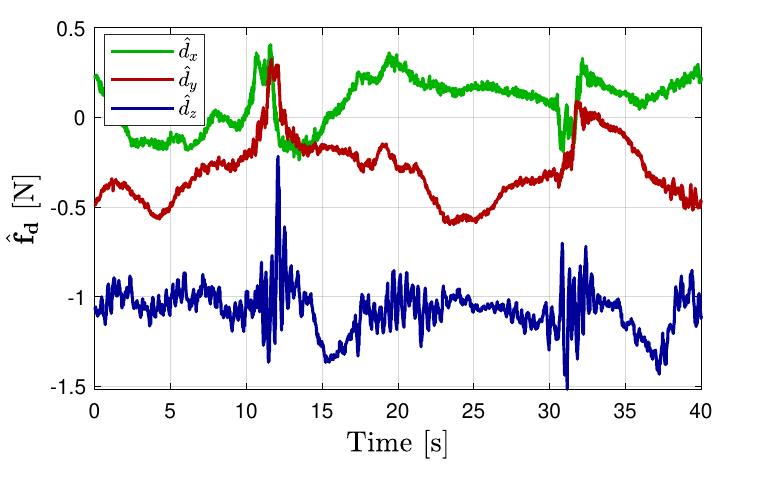}
    \caption{Disturbance estimates for the static-obstacle scenario under disturbance (hardware experiments)}
    \label{fig:d_hat}
    \centering
\end{figure}

To assess robustness under external disturbances, we compared the proposed IPC framework with and without HGDO in a static-obstacle scenario using two fans to generate wind. Figures \ref{fig:traj_real} and \ref{fig:u_real} present the resulting trajectory tracking and control signals. It is evident that, without HGDO, wind disturbances substantially degrade tracking, particularly near the points $(1,\,1.5)^\top$ and $(-1,\,1,0.5)^\top$. With HGDO, however, the vehicle effectively compensates for the disturbance, yielding markedly improved trajectory tracking. These results highlight the value of incorporating HGDO to achieve a robust IPC framework under external disturbances. Figure \ref{fig:d_hat} depicts the real-time disturbance estimates.

\section{Conclusion}
In this paper, we present an IPC framework for MRUAVs that embeds CBF terms as exponential penalties within NMPC, integrates an HGDO for disturbance compensation, and uses a KF for real-time prediction of moving-obstacle trajectories. Our approach addresses key limitations of prior work: standard NMPC often fails in dynamic settings because obstacle constraints ignore system dynamics, while NMPC with hard CBF constraints, though theoretically sound, can yield oscillatory, impractical motions under tight control bounds. By relaxing CBF constraints into tunable exponential penalties, our method preserves feasibility and improves practical safety by sharply increasing the penalty as the system approaches constraint boundaries. In extensive comparisons with NMPC baselines in Gazebo and on hardware, the proposed approach demonstrates superior feasibility, safety, and robustness. To the best of our knowledge, this is the first NMPC–CBF penalty IPC framework for MRUAVs validated in real flight, offering a practical IPC methodology for safe operation in confined, uncertain, and dynamic environments.

% Bibliography
\bibliographystyle{IEEEtran}
\bibliography{Reference}

@IEEEtranBSTCTL{IEEEexample:BSTcontrol,
	CTLuse_article_number     = "yes",
	CTLuse_paper              = "yes",
	CTLuse_forced_etal        = "yes",
	CTLmax_names_forced_etal  = "2",
	CTLnames_show_etal        = "1",
	CTLuse_alt_spacing        = "yes",
	CTLalt_stretch_factor     = "4",
	CTLdash_repeated_names    = "no",
	CTLname_format_string     = "{f.~}{vv~}{ll}{, jj}",
	CTLname_latex_cmd         = ""
}

@article{izadi2024hgdo,
  author = {Mohammadreza Izadi and Reza Faieghi},
  title = {High-gain disturbance observer for robust trajectory tracking of quadrotors},
  journal = {Control Engineering Practice},
  volume = {145},
  pages = {105854},
  year = {2024},
  publisher = {Elsevier},
  doi = {10.1016/j.conengprac.2024.105854}
}

@article{kamel2018review,
  title={A review on motion control of unmanned ground and aerial vehicles based on model predictive control techniques},
  author={Kamel, Mohamed A and Hafez, Ahmed T and Yu, Xiang},
  journal={Journal of Engineering Science and Military Technologies},
  volume={2},
  number={1},
  pages={10--23},
  year={2018},
  publisher={The Military Technical College}
}

@article{ames2016control,
  title={Control barrier function based quadratic programs for safety critical systems},
  author={Ames, Aaron D and Xu, Xiangru and Grizzle, Jessy W and Tabuada, Paulo},
  journal={IEEE Transactions on Automatic Control},
  volume={62},
  number={8},
  pages={3861--3876},
  year={2016},
  publisher={IEEE}
}

@article{bui2024model,
  title={Model Predictive Control for Optimal Motion Planning of Unmanned Aerial Vehicles},
  author={Bui, Duy-Nam and Khuat, Thu Hang and Phung, Manh Duong and Tran, Thuan-Hoang and Tran, Dong LT},
  journal={arXiv preprint arXiv:2410.09799},
  year={2024}
}

@inproceedings{small2019aerial,
  title={Aerial navigation in obstructed environments with embedded nonlinear model predictive control},
  author={Small, Elias and Sopasakis, Pantelis and Fresk, Emil and Patrinos, Panagiotis and Nikolakopoulos, George},
  booktitle={2019 18th European Control Conference (ECC)},
  pages={3556--3563},
  year={2019},
  organization={IEEE}
}

@article{luis2020online,
  title={Online trajectory generation with distributed model predictive control for multi-robot motion planning},
  author={Luis, Carlos E and Vukosavljev, Marijan and Schoellig, Angela P},
  journal={IEEE Robotics and Automation Letters},
  volume={5},
  number={2},
  pages={604--611},
  year={2020},
  publisher={IEEE}
}

@inproceedings{castillo2018model,
  title={Model predictive control for aerial collision avoidance in dynamic environments},
  author={Castillo-Lopez, Manuel and Sajadi-Alamdari, Seyed Amin and Sanchez-Lopez, Jose Luis and Olivares-Mendez, Miguel A and Voos, Holger},
  booktitle={2018 26th Mediterranean Conference on Control and Automation (MED)},
  pages={1--6},
  year={2018},
  organization={IEEE}
}

@article{ahn2022model,
  title={Model predictive control-based multirotor three-dimensional motion planning with point cloud obstacle},
  author={Ahn, Hyungjoo and Park, Junwoo and Bang, Hyochoong and Kim, Yoonsoo},
  journal={Journal of Aerospace Information Systems},
  volume={19},
  number={3},
  pages={179--193},
  year={2022},
  publisher={American Institute of Aeronautics and Astronautics}
}

@article{toumieh2024high,
  title={High-speed motion planning for aerial swarms in unknown and cluttered environments},
  author={Toumieh, Charbel and Floreano, Dario},
  journal={IEEE Transactions on Robotics},
  year={2024},
  publisher={IEEE}
}

@inproceedings{ali2024mpc,
  title={MPC based linear equivalence with control barrier functions for VTOL-UAVs},
  author={Ali, Ali Mohamed and Hashim, Hashim A and Shen, Chao},
  booktitle={2024 American Control Conference (ACC)},
  pages={1--6},
  year={2024},
  organization={IEEE}
}

@article{wang2024dual,
  title={Dual model predictive control of multiple quadrotors with formation maintenance and collision avoidance},
  author={Wang, Sifei and Wang, Yaonan and Miao, Zhiqiang and Wang, Xiangke and He, Wei},
  journal={IEEE Transactions on Industrial Electronics},
  year={2024},
  publisher={IEEE}
}

@article{goarin2024decentralized,
  title={Decentralized nonlinear model predictive control for safe collision avoidance in quadrotor teams with limited detection range},
  author={Goarin, Manohari and Li, Guanrui and Saviolo, Alessandro and Loianno, Giuseppe},
  journal={arXiv preprint arXiv:2409.17379},
  year={2024}
}

@article{romero2022model,
  title={Model predictive contouring control for time-optimal quadrotor flight},
  author={Romero, Angel and Sun, Sihao and Foehn, Philipp and Scaramuzza, Davide},
  journal={IEEE Transactions on Robotics},
  volume={38},
  number={6},
  pages={3340--3356},
  year={2022},
  publisher={IEEE}
}

@inproceedings{minavrik2024model,
  title={Model predictive path integral control for agile unmanned aerial vehicles},
  author={Mina{\v{r}}{\'\i}k, Michal and P{\v{e}}ni{\v{c}}ka, Robert and Von{\'a}sek, Vojt{\v{e}}ch and Saska, Martin},
  booktitle={2024 IEEE/RSJ International Conference on Intelligent Robots and Systems (IROS)},
  pages={13144--13151},
  year={2024},
  organization={IEEE}
}

@article{wu2021external,
  title={External forces resilient safe motion planning for quadrotor},
  author={Wu, Yuwei and Ding, Ziming and Xu, Chao and Gao, Fei},
  journal={IEEE Robotics and Automation Letters},
  volume={6},
  number={4},
  pages={8506--8513},
  year={2021},
  publisher={IEEE}
}

@article{liu2023integrated,
  title={Integrated planning and control for quadrotor navigation in presence of suddenly appearing objects and disturbances},
  author={Liu, Wenyi and Ren, Yunfan and Zhang, Fu},
  journal={IEEE Robotics and Automation Letters},
  volume={9},
  number={1},
  pages={899--906},
  year={2023},
  publisher={IEEE}
}

@inproceedings{ames2014control,
  title={Control barrier function based quadratic programs with application to adaptive cruise control},
  author={Ames, Aaron D and Grizzle, Jessy W and Tabuada, Paulo},
  booktitle={53rd IEEE conference on decision and control},
  pages={6271--6278},
  year={2014},
  organization={IEEE}
}

@inproceedings{nguyen2016exponential,
  title={Exponential control barrier functions for enforcing high relative-degree safety-critical constraints},
  author={Nguyen, Quan and Sreenath, Koushil},
  booktitle={2016 American Control Conference (ACC)},
  pages={322--328},
  year={2016},
  organization={IEEE}
}

@article{tajeddin2019ecological,
  title={Ecological adaptive cruise control with optimal lane selection in connected vehicle environments},
  author={Tajeddin, Sadegh and Ekhtiari, Sanaz and Faieghi, Mohammadreza and Azad, Nasser L},
  journal={IEEE Transactions on Intelligent Transportation Systems},
  volume={21},
  number={11},
  pages={4538--4549},
  year={2019},
  publisher={IEEE}
}

@article{xiao2021adaptive,
  title={Adaptive control barrier functions},
  author={Xiao, Wei and Belta, Calin and Cassandras, Christos G},
  journal={IEEE Transactions on Automatic Control},
  volume={67},
  number={5},
  pages={2267--2281},
  year={2021},
  publisher={IEEE}
}

@article{shayan2025exponential,
  title={Exponential control barrier function and model predictive control for jerk-level reactive motion planning of quadrotors},
  author={Shayan, Zeinab and Izadi, Mohammadreza and Scognamiglio, Vincenzo and D’Angelo, Simone and Singoji, Shashank and Lippiello, Vincenzo and Faieghi, Reza},
  journal={Control Engineering Practice},
  volume={164},
  pages={106489},
  year={2025},
  publisher={Elsevier}
}

@article{verschueren2022acados,
  title={acados—a modular open-source framework for fast embedded optimal control},
  author={Verschueren, Robin and Frison, Gianluca and Kouzoupis, Dimitris and Frey, Jonathan and Duijkeren, Niels van and Zanelli, Andrea and Novoselnik, Branimir and Albin, Thivaharan and Quirynen, Rien and Diehl, Moritz},
  journal={Mathematical Programming Computation},
  volume={14},
  number={1},
  pages={147--183},
  year={2022},
  publisher={Springer}
}

@article{he2015stability,
  title={On stability of multiobjective NMPC with objective prioritization},
  author={He, Defeng and Wang, Lei and Sun, Jing},
  journal={Automatica},
  volume={57},
  pages={189--198},
  year={2015},
  publisher={Elsevier}
}

@inproceedings{Zeilinger2014,
  title={Robust stability properties of soft constrained MPC},
  author={Zeilinger, Melanie N and Jones, Colin N and Morari, Manfred},
  booktitle={49th IEEE Conference on Decision and Control (CDC)},
  pages={5276--5282},
  year={2010},
  organization={IEEE}
}

@article{shayan2025nonlinear,
  title={Nonlinear model predictive control of tiltrotor quadrotors using feasible control allocation},
  author={Shayan, Zeinab and Cristobal, Jann and Izadi, Mohammadreza and Yazdanshenas, Amin and Naderi, Mehdi and Faieghi, Reza},
  journal={Journal of Intelligent \& Robotic Systems},
  volume={111},
  number={2},
  pages={54},
  year={2025},
  publisher={Springer}
}

\balance
\end{document}